# A Time-Enhanced Data Disentanglement Network for Traffic Flow Forecasting

Tianfan Jiang[1][0009-0001-9880-1556], Mei Wu[1][0009-0007-3395-2891], Wenchao Weng[3][0000-0002-9677-3213], Dewen Seng[2✉][0009-0001-9880-1556], and Yiqian Lin[2][0009-0000-4306-3392]

[1] ITMO Joint Insistute, Hangzhou Dianzi University, Hangzhou 310018, China
[2] School of Computer Science and Technology, Hangzhou Dianzi University, Hangzhou 310018, China
`{232320005, 222320007, sengdw, linyq}@hdu.edu.cn`
[3] College of Computer Science and Technology, Zhejiang University of Technology, Hangzhou 310023, China
`111124120010@zjut.edu.cn`

**Abstract.** In recent years, traffic flow prediction has become a highlight in the field of intelligent transportation systems. However, due to the temporal variations and dynamic spatial correlations of traffic data, traffic prediction remains highly challenging.Traditional spatiotemporal networks, which rely on end-to-end training, often struggle to handle the diverse data dependencies of multiple traffic flow patterns. Additionally, traffic flow variations are highly sensitive to temporal information changes. Regrettably, other researchers have not sufficiently recognized the importance of temporal information.To address these challenges, we propose a novel approach called A Time-Enhanced Data Disentanglement Network for Traffic Flow Forecasting (TEDDN). This network disentangles the originally complex and intertwined traffic data into stable patterns and trends. By flexibly learning temporal and node information through a dynamic graph enhanced by a temporal feature extraction module, TEDDN demonstrates significant efficacy in disentangling and extracting complex traffic information. Experimental evaluations and ablation studies on four real-world datasets validate the superiority of our method.

**Keywords:** Traffic Flow Prediction, Traffic Data Disentanglement, Temporal Enhancement, Dynamic Graph Learning.

## 1 Introduction

Accurate traffic flow forecasting is crucial for urban transportation planning, management, and the development of intelligent transportation systems. It can provide references for infrastructure layout, facilitate the optimization of traffic management, and promote the progress of intelligent transportation technologies [1]. It involves using historical traffic observation data to predict the inflow and outflow of vehicles in specific areas at given times. During the recent years, a significant shift has occurred towards harnessing deep learning techniques to adeptly detect and analyze the intricate spatiotemporal relationships embedded within traffic data. Within this domain, adaptive spatial-temporal graph neural networks have particularly stood out as a key area of focus [2]. Owing to their ability to delineate the complex interconnections and dependencies between entities within non-Euclidean frameworks, these networks have achieved notable efficacy in the realm of traffic flow forecasting. Conventional methods for predicting traffic flow often hinge on raw historical traffic data, which may overlook the intricate and intertwined relationships in modern traffic systems. It is well known that different traffic modes can interfere with each other, thereby increasing the dynamic complexity of traffic systems and limiting the precision of traffic flow forecasts.

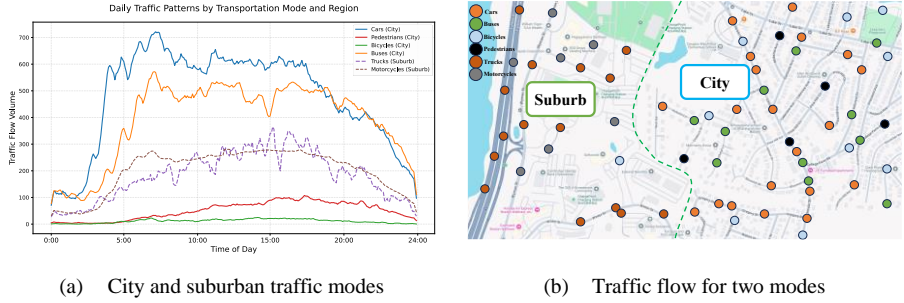

(a)   City and suburban traffic modes       (b)   Traffic flow for two modes

**Fig. 1.**   Two traffic patterns between ciyies and suburbs

In our daily lives, people always utilize various modes of transportation, thereby forming diverse traffic flow patterns. Fig. 1 clearly demonstrate the flow characteristics generated by different transportation vehicles over the past 24 hours. Distinct transportation modes exhibit markedly different flow patterns: urban roads dominated by cars and buses show continuously increasing traffic flow from midnight until after 8 AM, followed by a sustained decline from 11 AM to 3 PM, while consistently maintaining relatively high traffic volume. In contrast, suburban areas (indicated by dashed lines in the figures) display unique flow patterns for trucks and motorcycles, with their traffic flow steadily rising from 5 AM until peaking in the early afternoon, followed by a gradual decline. However, current research often over-relies on single-mode traffic information for traffic prediction, neglecting the fact that traffic data essentially represents the integrated outcome of multiple transportation modes interacting and influencing each other.

To effectively handle traffic data entangled with multiple traffic modes and better extract crucial temporal information for traffic flow, we propose a novel traffic flow prediction network: A Time-Enhanced Data Disentanglement Network for Traffic Flow Forecasting (TEDDN). To capture its spatiotemporal features, TEDDN preprocesses historical traffic data using temporal and node embeddings. Through the Time-Enhanced Module (TE Module), the distilled temporal information is enriched in its representational capacity and filtered for noise [3]. The historical data is then disentangled into temporal and spatial features, with each resulting dataset representing a specific traffic mode. Additionally, the graph learning layer generates a dynamic, learnable graph that captures dynamic spatial dependencies using an integrated gated graph convolution and data disentanglement module [4]. Outputs from various elements are merged over time to grasp varied-scale depictions. Our primary contributions are encapsulated as follows:

- We propose a novel approach to traffic data processing, termed the Disentangle Gate. This method involves segregating data across various transportation modes to uncover intrinsic correlations between traffic flow and these modes. By doing so, we can more effectively learn the unique characteristics of each mode, thereby enhancing the accuracy and efficiency of predictions. A series of ablation studies have been conducted, confirming the significant advantages of our disentanglement module in practical applications.
- We introduce the Time-Enhanced Module (TE Module), which generates an attention coefficient in each temporal information cluster to adjust sub-feature significance. This enhances each cluster's representation and reduces noise, optimizing the signal-to-noise ratio. The attention coefficients are determined by the affinity between global and local feature descriptors, keeping the module streamlined and computationally efficient.
- We perform thorough experiments across four real-world, large-scale datasets to fully probe the TEDDN framework and verify its efficacy. The consistent experimental results consistently show a marked superiority over all benchmark approaches.

## 2    Related Work

### 2.1    Traffic Forecasting

Researchers initially used linear statistical methods for traffic prediction, such as historical averages and autoregressive models, but these were limited in capturing the nonlinear dynamics of traffic data [5]. The advent of machine learning introduced techniques like support vector regression and K-nearest neighbors [6], which, despite their limitations in feature generalization, marked a step forward. Deep learning's rise brought sophisticated models like LSTM [7], TCN [8], and Transformer to individually analyze sensor data for temporal patterns. However, these models often disregarded the spatial interdependencies within traffic networks. To address this, hybrid models combining GCNs with sequential methods, such as STGCN [9] and DCRNN [10], were developed to capture spatiotemporal correlations. Advanced models like GWN and AGCRN further refined this by using adaptive graphs for dynamic spatial dependency capture, though they risked overfitting or underfitting due to the lack of prior knowledge. STFGNN proposes a data-driven graph structure based on the Dynamic Time Warping (DTW) algorithm [11]. STFGNN and STGODE models, leveraging spatiotemporal fusion graphs and tensor-based neural ODEs, respectively, have shown promise in improving accuracy while mitigating oversmoothing. Despite these advancements, most GCN-based methods still do not fully account for the dynamic and diverse nature of traffic patterns across the road network, indicating a need for further research in this area.

### 2.2    Spatial-Temporal Graph Neural Networks

Recent research has underscored the importance of spatio-temporal modeling techniques, particularly in intelligent transportation systems. These techniques combine temporal convolutional networks with graph structures to capture how node features change over time, thus improving the accuracy of urban traffic flow predictions. Spatio-Temporal Graph Convolutional Networks [12] integrate graph convolutional networks and temporal learning methods to reveal the latent relationships between temporal dynamics and spatial context. Early studies, like DCRNN, used bidirectional random walks and an encoder-decoder framework for spatial and temporal modeling. As research has progressed, Adaptive Spatio-Temporal Graph Neural Networks have become a new trend [13]. Examples include Graph WaveNet's adaptive graph convolution layer, STFGNN's data-driven graph structure based on the DTW algorithm, and DDGCRN's combination of dynamic graph convolution and RNN architecture to respond to time-varying traffic signals [14]. By directly analyzing historical traffic flow data, the Dynamic Spatial-Temporal Aware Graph Neural NetworkMadeptly captures the dynamic nature of spatial associations between nodes, thereby extracting both temporal and spatial correlations. Additionally, the incorporation of reinforcement learning for generating dynamic graphs significantly enhances the accuracy and efficiency of spatio-temporal feature extraction, markedly improving the precision of urban traffic data modeling and prediction [15].

## 3    Preliminary

We embark on an exploration of the fundamental constructs of traffic networks and signals, delineating the theoretical underpinnings that govern their operation.

**Definition 1** (Traffic Graph). The graph is defined as $G = (V, E)$ where $V$ denotes the vertex set, and $E$ denotes the edge set. The cardinality of the vertex set in a graph is indicated by $N$. The connectivity between vertices is encapsulated in an adjacency matrix $A \in \mathbb{R}^{N \times N}$.

**Definition 2** (Traffic Signal). We depict the traffic inflow and outflow data from the preceding $T$ time steps via a traffic tensor. $X_{t-T_h:t} \in \mathbb{R}^{(T_h \times N \times C)} = (X_{t-T_h-1}, \ldots, X_t)$. The traffic volume information at the t-$t_h$ time slot is denoted as $X_t \in \mathbb{R}^{(N \times C)}$, where $C$ is the number of features collected by the sensors.

## 4    Methodology

In this section, we will provide a detailed overview of the framework we propose, as shown in Fig. 2.

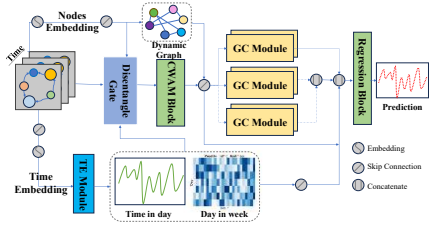
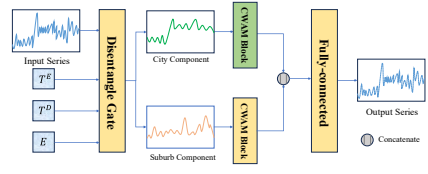

**Fig. 2.** The framework of TEDDN       **Fig. 3.** Disentangle Gate

### 4.1 Overview

The pipeline first embeds temporal and node information to extract spatiotemporal features. At step $t$, the TE Module refines temporal hidden states ($T_t^D$, $T_t^W$) as supplementary inputs. Simultaneously, a graph learning layer generates dynamic structures processed by Disentangle Gate with channel squeeze. The disentangled features are concatenated and fed into the GC Module for traffic mode differentiation. Finally, time-embedded hidden states are integrated into auxiliary inputs, while the output layer projects features via parameter matrix $W$ for prediction.

### 4.2 CWAM Block

We propose a method named Channel-Wise Attention Mechanism (CWAM) for condensing global spatial information into a channel descriptor. The statistic $z \in \mathbb{R}^C$ is created by reducing the spatial dimensions $H \times W$ of the temporal information features, where the $c$-$t_h$ element of $z$ is calculated as follows:

$$z_c = GlobalAveragePooling(u_c) \quad (1)$$

We select to implement a straightforward gating mechanism, employing a sigmoid activation function.

$$s = Sigmoid(g(z, W)) = Sigmoid(W_2 ReLU(W_1 z)) \quad (2)$$

Here, $s$ denotes the activation output, $W, W_1, W_2$ represent weight matrices, and $z$ is the aggregated statistic. To enhance generalization while controlling complexity, we employ a parameterized gating mechanism.

$$(\tilde{x}_c) = F_{scale}(u_c, s_c) = s_c u_c \quad (3)$$

In this context, $F_{scale}(\cdot)$ denotes the channel multiplication function, $s_c$ denotes the resultant of the activation function for each channel, and the processed feature set $\tilde{X} = [\tilde{x}_1, \tilde{x}_2, \tilde{x}_3, \dots, \tilde{x}_C]$ is obtained through the channel features $(\tilde{x}_C)$.

### 4.3 Disentangle Gate

We use two randomly initialized, learnable time slot embedding matrices $T^D \in \mathbb{R}^{N_D \times d}$ and $T^W \in \mathbb{R}^{N_W \times d}$ to convert date and time attributes into continuous vectors, encoding temporal information. The node embedding matrix $E$ captures semantic information and features of nodes in the traffic network. Using a CWAM Block, we optimize feature weights to prioritize significant information. To estimate proportional relationships between hidden traffic flows, we apply a Sigmoid activation function to generate a decouple gate value. The disentangled gate value is then multiplied step by step with the unprocessed raw traffic flow data to distinguish and categorize different traffic patterns. The detailed schematic is shown in Fig. 3.

$$\Omega_{t,i} = Sigmoid(ReLU(T_t^D||(T_t^W||E_i)W_1)W_2) \tag{4}$$

$$X_1 = X \odot \Omega \tag{5}$$

$$X_2 = X - X_1 \tag{6}$$

The weight matrices $W_1 \in \mathbb{R}^{(2D+Nd)\times d}$ and $W_2 \in \mathbb{R}^{d\times 1}$ are used to apply linear transformations to the frature vectors. Additionally, $\Omega_{t,i} \in (0,1)$ indicates the ratio of the traffic flow pattern associated with node $j$ at time step $t$ to the total traffic flow on the road at that time. The dimension of matrix $\Omega$ is $\mathbb{R}^{Th\times N\times 1}$. Multiplying the disentangle gate with the original traffic flow yield $X_1$, representing a specific traffic pattern's contribution. Subtracting $X_1$ from $X$ gives $X_2$, which includes the residual flow not explained by the specific traffic pattern.

### 4.4    TE Module

We use holistic group time data to optimize feature learning in key areas. The global statistical feature $F_{gp}$ of the spatial averaging function can help estimate the temporal information vector for group learning.

$$g = F_{gp}(X) = Average(x_i) \tag{7}$$

Where the spatial averaging function $F_{gp}(X)$ is applied to the global temporal feature $X$ to obtain the global temporal vector $g$, and the dot product is used to calculate the importance coefficient:

$$c_i = g \odot x_i \tag{8}$$

Where $x_i$ represents the temporal feature of each node, and a stability constant $\varepsilon$ is introduced in the spatial domain of the importance coefficient:

$$\hat{c}_i = \frac{c_i - \mu_c}{\sigma_c + \epsilon} \tag{9}$$

$$\mu_c = \frac{1}{m}\sum_{j=1}^{m} c_j \tag{10}$$

$$\sigma_c^2 = \frac{1}{m}\sum_{j=1}^{m}(c_j - \mu_c)^2 \tag{11}$$

$\mu_c$ represents the mean of the $m$ importance coefficients $c_j$, and $\sigma_c^2$ represents the variance of the $m$ importance coefficients $c_j$. Introducing the stability constant $\epsilon$ into the importance coefficient $\hat{c}_i$, and introducing normalization parameters $\gamma$ and $\beta$ that reflect identity transformation, scaling, and shifting, the final importance coefficient is obtained:

$$a_i = \gamma \hat{c}_i + \beta \qquad (12)$$

Using an S-shaped gating mechanism, the importance coefficient $a_i$ is used to modulate the feature $x_i$, resulting in the enhanced feature:

$$\hat{x}_i = x_i \odot Sigmoid(a_i) \qquad (13)$$

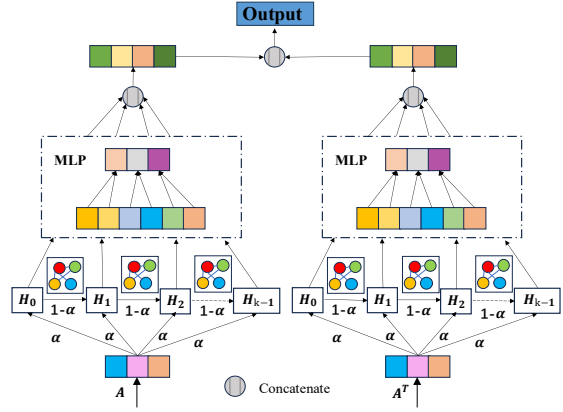

**Fig. 4.** GC Module

### 4.5 GC Module

This module precisely models complex node-edge relationships in traffic networks to effectively analyze spatiotemporal dependencies (see Fig. 4 for architecture). The non-negative adjacency matrix is constructed as:

$$DE_1 = \tanh(\alpha E_1 W_1) \qquad (14)$$

$$DE_2 = \tanh(\alpha E_2 W_2) \qquad (15)$$

$$A = ReLU(\tanh(\alpha(DE_1 DE_2^T - DE_1^T DE_2))) \qquad (16)$$

Where $E_1$ and $E_2$ represent the embedding matrices of two traffic network nodes, $W_1$ and $W_2$ represent the corresponding weight matrices, $\alpha$ is a scalar used for scaling the matrices, $DE_1$ and $DE_2$ represent the intermediate results after linear transformation and nonlinear activation of the embedding matrices. Subtracting $DE_2$ from $DE_1$ forms the adjacency matrix $\tilde{A}$, and applying the ReLU activation function produces the non-negative adjacency matrix $A$. The expression for the residual graph convolution layer is as follows:

$$H^{(k)} = \beta H_{in} + (1 - \beta)(\tilde{D}^{-1}\tilde{A})H^{(k-1)} \qquad (17)$$

$$H_{out} = (H^{(0)}||H^{(1)}||...||H^{(k-1)})W \quad (18)$$

The residual graph convolution layer processes hidden input $H_{in}$ to generate same-dimensional state $H^{(k)}$ (propagation depth $k$), where $\beta$ controls initial state retention, $\widetilde{D}$ is the degree matrix, $W$ denotes dense layer parameters, and $H_{out}$ represents the final output tensor.

## 5 Experiments

### 5.1 Datasets

We evaluated TEDDN using Caltrans' PeMS datasets (PEMS03-08) with 5-minute aggregated traffic data, which have been widely adopted in previous studies [16]. The model predicts 60-minute traffic flow (12 steps) from historical 60-minute data. See Table 1 for dataset details.

**Table 1.** Summary of Datasets.

| Datasets | #Nodes | #Rate | #Time Range |
|---|---|---|---|
| PEMS03 | 358 | 26209 | 05/2012-07/2012 |
| PEMS04 | 307 | 16992 | 01/2018-02/2018 |
| PEMS07 | 883 | 28224 | 05/2017-08/2017 |
| PEMS08 | 170 | 17856 | 07/2016-08/2016 |

### 5.2 Baseline Methods

**Baseline Methods:** FC-LSTM [17], DSANet [18], GraphwaveNet [19], DCRNN, ASTGCN [20], STFGNN [21], STGODE [22], STG-NCDE [23], DSTAGNN [24], ST-AE [25], PDFormer [26].

### 5.3 Experiment Settings

Experiments were executed on an NVIDIA GeForce RTX 4090 GPU. We split the PEMS03, PEMS04, PEMS07, and PEMS08 datasets into training, validation, and test sets in a 6:2:2 ratio. The Adam optimizer was used for training with a weight decay of $1.0 \times 10^{-5}$ and epsilon of $1.0 \times 10^{-8}$. A learning rate scheduler with a decay rate of 0.5 was applied, along with a 30-epoch warm-up period and curriculum learning, incrementing by one every three epochs up to 12 steps. An early stopping mechanism with a patience of 100 epochs was implemented to guard against overfitting. The batch size is 32, and the learning rate was set to 0.002. Model performance was assessed using Mean Absolute Error (MAE), Mean Absolute Percentage Error (MAPE), and Root Mean Square Error (RMSE).

Table 2. Performance on PEMS03, 04, 07, and 08.

| Models | PEMS03 | | | PEMS04 | | | PEMS07 | | | PEMS08 | | |
|---|---|---|---|---|---|---|---|---|---|---|---|---|
| | MAE | RMSE | MAPE | MAE | RMSE | MAPE | MAE | RMSE | MAPE | MAE | RMSE | MAPE |
| FC-LSTM | 21.33 | 35.11 | 23.33% | 27.14 | 41.59 | 18.20% | 29.98 | 45.94 | 13.20% | 22.20 | 34.06 | 14.20% |
| DSANet | 21.29 | 34.55 | 23.21% | 22.79 | 35.77 | 16.03% | 32.36 | 49.11 | 14.43% | 17.14 | 26.96 | 11.32% |
| GraphWave-Net | 19.12 | 32.77 | 18.89% | 24.89 | 39.66 | 17.29% | 26.39 | 41.50 | 11.97% | 18.28 | 30.05 | 12.15% |
| DCRNN | 18.30 | 29.74 | 17.86% | 23.54 | 36.25 | 17.18% | 23.87 | 37.27 | 10.50% | 18.41 | 28.28 | 12.17% |
| ASTGCN | 17.34 | 29.56 | 17.21% | 22.92 | 35.22 | 16.56% | 24.01 | 37.87 | 10.73% | 18.25 | 28.06 | 11.64% |
| STFGNN | 16.77 | 28.34 | 16.30% | 19.83 | 31.88 | 13.02% | 22.07 | 35.80 | 9.21% | 16.64 | 26.22 | 10.60% |
| STGODE | 16.50 | 27.84 | 16.69% | 20.84 | 32.82 | 13.77% | 22.59 | 37.54 | 10.14% | 16.81 | 25.97 | 10.62% |
| STG-NCDE | 15.57 | 27.09 | 15.06% | 19.21 | 31.09 | 12.76% | 20.53 | 33.84 | 8.80% | 15.54 | 24.81 | 9.92% |
| DSTAGNN | 15.57 | 27.21 | 14.68% | 19.30 | 31.46 | 12.70% | 21.42 | 34.51 | 9.01% | 15.67 | 24.77 | 9.94% |
| ST-AE | 15.44 | 26.29 | 15.20% | 19.74 | 31.14 | 15.40% | 22.75 | 35.56 | 10.18% | 15.96 | 25.03 | 11.34% |
| PDFormer | 14.94 | 25.39 | 15.82% | 18.37 | 30.03 | **12.10**% | 19.83 | 32.87 | 8.53% | **13.58** | 23.50 | **9.04**% |
| TEDDN | **14.70** | **24.35** | **14.39**% | **18.31** | **30.02** | 12.17% | **19.56** | **32.78** | **8.15**% | 13.70 | **23.11** | 9.17% |

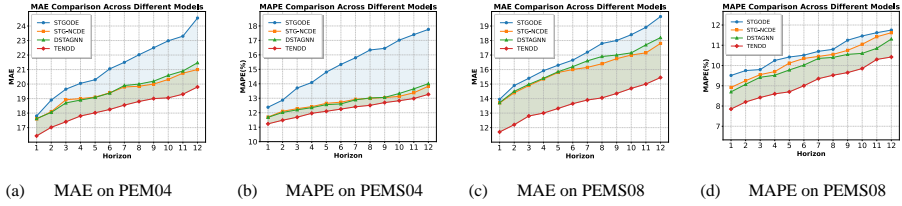

(a)   MAE on PEM04    (b)   MAPE on PEMS04    (c)   MAE on PEMS08    (d)   MAPE on PEMS08

Fig. 5. Prediction error at each time horizon for baseline models

### 5.4   Experiment Results

Experimental results in Table 2 show TEDDN outperforms mainstream traffic prediction models across four datasets, demonstrating superior performance in capturing traffic flow dynamics. While non-graph models like FC-LSTM handle temporal features well but struggle with spatial variations, and graph-based approaches like DCRNN better capture spatial correlations yet still face limitations in pattern recognition, TEDDN's data disentanglement mechanism proves critical for processing complex traffic data. Notably, TEDDN achieves the best overall performance, maintaining stable accuracy across all datasets compared to DSTAGNN's strength on PEMS03/PEMS04 and STG-NCDE's advantage on PEMS04/PEMS08. As shown in Fig. 5, TEDDN's consistent performance across both short- and long-term predictions contrasts with STGODE's faster error accumulation in single-step forecasts, confirming TEDDN's robustness.

### 5.5   Ablation Experiments

To validate the effectiveness of different components in TEDDN, we conducted a series of ablation experiments on the PEMS08 dataset. These experiments encompassed four distinct variants of the model, each designed to assess the contribution of specific components to the overall performance of TEDDN. By systematically removing or modifying individual components, we aimed to gain insights into their respective impacts on the model's predictive accuracy and robustness.

- **w/o TE:** Remove the TE Module.
- **w/o DG:** Remove the Disentangle Gate.
- **w/o GRU:** Remove the GRU layer.
- **TEDDN:** Complete model.

Table 3. Ablation Study on PEMS08.

| Variants | Horizon 3 | | | Horizon 6 | | | Horizon 12 | | | Average | | |
|---|---|---|---|---|---|---|---|---|---|---|---|---|
| | MAE | RMSE | MAPE | MAE | RMSE | MAPE | MAE | RMSE | MAPE | MAE | RMSE | MAPE |
| w/o TE | 13.17 | 21.53 | 8.46% | 13.87 | 23.39 | 9.14% | 15.49 | 25.70 | 10.26% | 13.89 | 34.06 | 14.20% |
| w/o DG | 13.95 | 22.26 | 9.67% | 14.99 | 24.22 | 10.25% | 17.05 | 27.20 | 11.95% | 15.06 | 26.96 | 11.32% |
| w/o GRU | 13.09 | 21.58 | 8.52% | 14.02 | 23.44 | 9.15% | 15.59 | 25.82 | 10.61% | 14.06 | 30.05 | 12.15% |
| **TEDDN** | **12.68** | **21.27** | **8.39%** | **13.60** | **23.08** | **9.00%** | **15.37** | **25.68** | **10.43%** | **13.70** | **23.11** | **9.17%** |

As shown in Table 3, TEDDN outperforms w/o TE at the 3rd horizon, 6th horizon, and 12th horizon, indicating that the temporal information enhancement module improves the model's ability to handle temporal information and plays a positive role in the framework, highlighting the importance of temporal information for traffic flow prediction. Additionally, we observe that the performance of w/o DG is significantly inferior to TEDDN, demonstrating that the disentangle Gate effectively disentangles complex traffic flow information and processes different flows specifically, playing a crucial role in the model. Lastly, w/o GRU consistently underperforms TEDDN across all time steps, indicating that both short-term and long-term dependencies are essential for accurate traffic forecasting.Through these experimental results, we can further understand the specific roles of each component in the model and provide valuable references for future research.

Table 4. Processing time on PEMS04 and PEMS08.

| Dataset | Variants | Training(s/epochs) | Inference(s) |
|---|---|---|---|
| PEMS04(32) | w/o TE | 58.76 | 7.11 |
| | w/o DG | 63.97 | 7.26 |
| | w/o GRU | 60.31 | 7.14 |
| | TEDDN | 65.48 | 7.43 |
| PEMS08(32) | w/o TE | 18.34 | 2.89 |
| | w/o DG | 18.98 | 2.74 |
| | w/o GRU | 19.76 | 2.79 |
| | **TEDDN** | **21.35** | **3.11** |

Table 4 presents the processing time required for different variants formed by various key components of our model on the PEMS04 and PEMS08 datasets, with a batch size set to 32. It can be observed that our TE Module, Disentangle Gate, and GRU gating mechanism do not impose a significant burden on the model's processing time, which demonstrates the efficiency of our model and its low complexity.

## 6  Conclusion

This paper proposes TEDDN, a traffic flow prediction model that employs temporal enhancement modules for feature extraction, disentangles traffic flows using spatio-temporal features, and processes them via adaptive graph residual convolution. Experiments show significant performance improvements across multiple datasets, with ablation studies confirming the efficacy of both the disentanglement and temporal modules without substantially increasing model complexity.

**Acknowledgments.** This research was funded by the Ministry of Education's Industry-University Cooperative Education Project under Grant No. 231106093150720 and 220903242265640.